\documentclass{article}

\usepackage{arxiv}

\usepackage{amsmath,amsfonts,bm}

\def\eqref#1{equation~\ref{#1}}

\def\1{\bm{1}}

\DeclareMathAlphabet{\mathsfit}{\encodingdefault}{\sfdefault}{m}{sl}
\SetMathAlphabet{\mathsfit}{bold}{\encodingdefault}{\sfdefault}{bx}{n}

\usepackage{tabularx}
\usepackage{wrapfig}
\usepackage{natbib}
\RequirePackage{algorithm}
\RequirePackage{algorithmic}
\usepackage{microtype}
\usepackage{amsmath}
\usepackage{amssymb}
\usepackage{mathtools}
\usepackage{amsthm}
\usepackage{url}
\usepackage{amsthm}
\usepackage{authblk}
\usepackage{microtype}
\usepackage{graphicx}
\usepackage{subfigure}
\usepackage{booktabs} 
\usepackage{multirow}
\usepackage{fancyvrb}
\usepackage{fvextra}
\usepackage[pagebackref,breaklinks,colorlinks,allcolors=myblue]{hyperref}

\usepackage[capitalize,noabbrev]{cleveref}
\usepackage{makecell}
\usepackage[table]{xcolor}
\usepackage{xcolor}
\definecolor{myblue}{rgb}{0.21,0.49,0.74}
\definecolor{MyGreen}{RGB}{50, 205, 50}  
\definecolor{lightblue}{RGB}{230,240,255}
\usepackage{enumitem}
\usepackage{subcaption}
\usepackage{array}
\usepackage{graphicx}
\usepackage{times}
\usepackage{latexsym}
\usepackage[most]{tcolorbox}  
\newtcolorbox{methodbox}[1]{
    enhanced,
    title={#1},  
    colback=gray!10,  
    colframe=black,   
    coltitle=white,   
    colbacktitle=black, 
    fonttitle=\bfseries,
    attach boxed title to top center={yshift*=0mm}, 
    boxed title style={
        size=small,
        colback=black,
        width=\textwidth, 
        center title,     
    },
    sharp corners,
    width=\textwidth,
    top=2em 
}

\usepackage[T1]{fontenc}
\usepackage{colortbl}

\usepackage[utf8]{inputenc}

\usepackage{microtype}

\usepackage{inconsolata}

\usepackage{graphicx}

\title{Temporal Consistency for LLM Reasoning Process Error Identification}

\author[1]{Jiacheng Guo}
\author[2]{Yue Wu}
\author[1]{Jiahao Qiu}
\author[1]{Kaixuan Huang}
\author[3]{Xinzhe Juan}
\author[1,2]{Ling Yang}
\author[1]{Mengdi Wang}
\affil[1]{Department of Electrical \& Computer Engineering, Princeton University}
\affil[2]{AI Lab, Princeton University} 
\affil[3]{Department of Computer Science \& Engineering, University of Michigan}

\theoremstyle{plain}

\theoremstyle{definition}

\theoremstyle{remark}

\begin{document}
\maketitle
\begin{abstract}
Verification is crucial for effective mathematical reasoning. We present a new \textit{temporal consistency} method where verifiers iteratively refine their judgments based on the previous assessment. Unlike one-round verification or multi-model debate approaches, our method leverages consistency in a sequence of self-reflection actions to improve verification accuracy. Empirical evaluations across diverse mathematical process error identification benchmarks (Mathcheck, ProcessBench, and PRM800K) show consistent performance improvements over baseline methods. When applied to the recent DeepSeek R1 distilled models, our method demonstrates strong performance, enabling 7B/8B distilled models to outperform all 70B/72B models and GPT-4o on ProcessBench. Notably, the distilled 14B model with our method achieves performance comparable to Deepseek-R1. Our codes are available at \url{https://github.com/jcguo123/Temporal-Consistency}
\end{abstract}
\begin{figure*}[h]
    \centering
    \vspace{-0.1in}
    \includegraphics[width=1.0\linewidth]{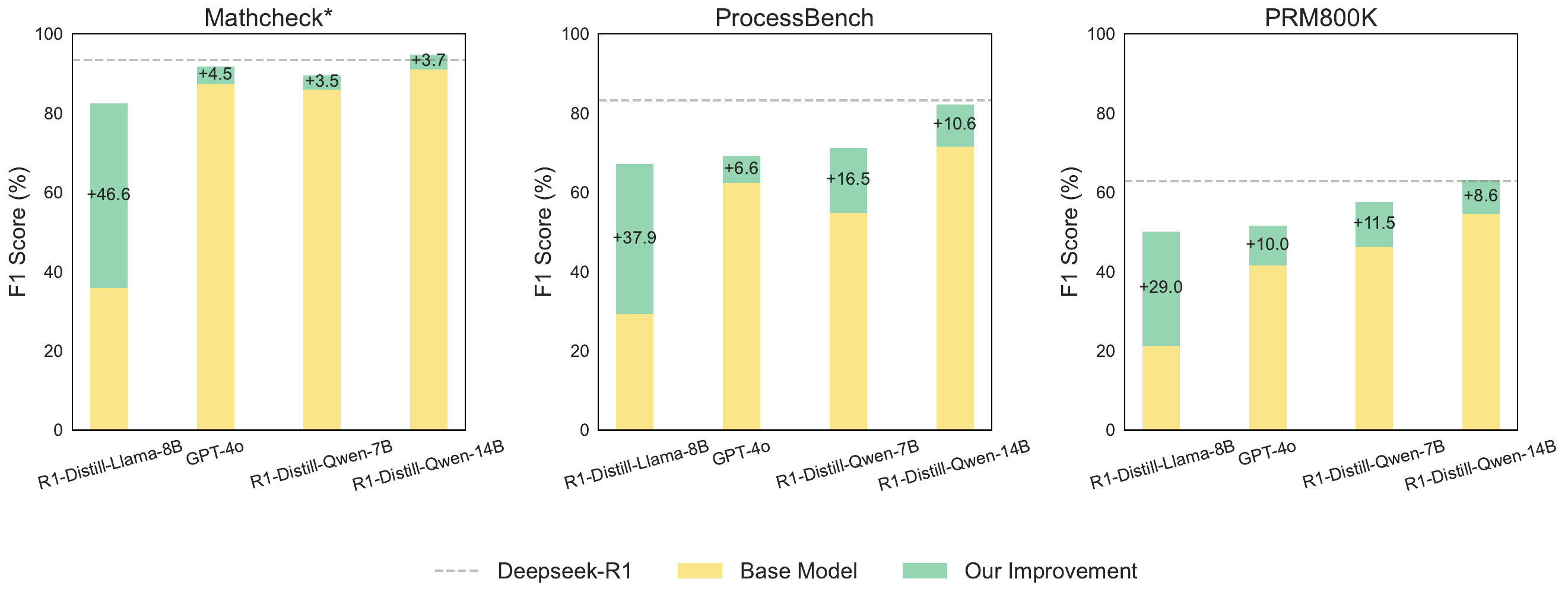}
    \caption{Performance improvements for various models on process error identification benchmarks.}
    \label{fig:0}
    \vspace{-0.1in}
\end{figure*}
\section{Introduction}
Large language models (LLMs) have shown impressive capabilities in reasoning tasks~\citep{grattafiori2024llama,yang2024qwen25mathtechnicalreportmathematical,jaech2024openai,guo2025deepseek,yang2025reasonflux}, but still often make mistakes when generating complex multi-step solutions. To address this issue, Process Reward Models (PRMs) \citep{lightman2023let,luo2024improve} have been introduced to guide generations. Instead of providing feedback solely on the final answer, PRMs evaluate every intermediate step in the reasoning chain, thereby aligning the model's chain of thought with correct logical sequences.

However, existing PRMs face several key limitations that hinder their broader applicability. First, training a PRM requires large-scale, high-quality annotated datasets, making the process highly data-intensive and costly to scale \citep{guo2025deepseek}. Second, PRMs exhibit poor out-of-domain generalization; models trained on specific problem distributions often struggle to accurately evaluate reasoning steps when confronted with diverse problem types \citep{zeng2025versaprmmultidomainprocessreward, lin2024limitedgeneralizationcapabilityimplicit}. Finally, the effectiveness of PRMs is intrinsically limited by the capability of the base model \citep{luo2024improvemathematicalreasoninglanguage}. These challenges highlight the need for further research to develop more scalable process supervision techniques in LLMs.

\begin{figure*}[t]
    \centering
    \includegraphics[width=1.0\textwidth]{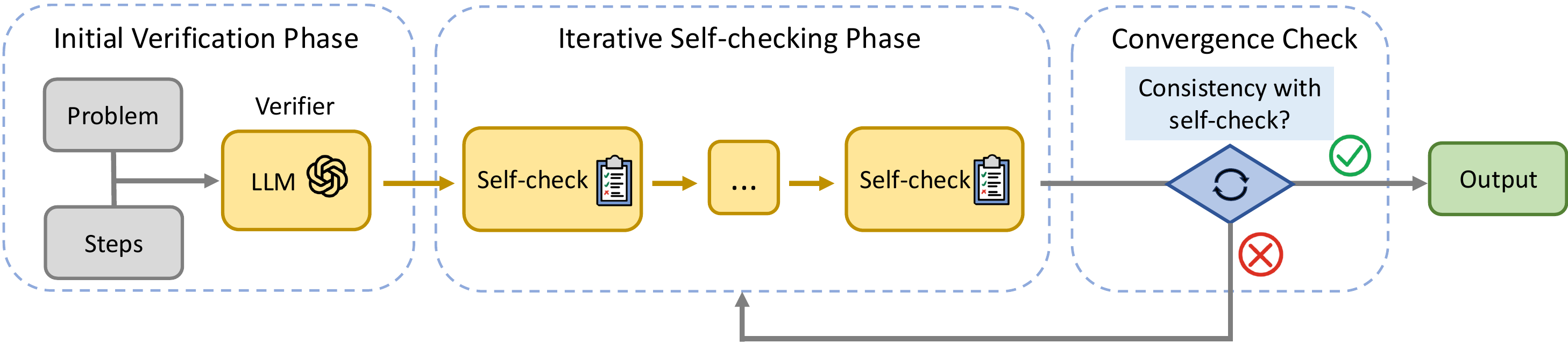}
    \caption{Overview of our Temporal Consistency approach, where each LLM iteratively examines its own verification results until reaching a stable result (stopping criteria defined in Section \ref{algorithm}). The self-checking mechanism allows LLMs to refine their judgments based on previous verifications, potentially correcting initial misidentification. }
    \vspace{-0.2in}
    \label{fig:1}
\end{figure*}
An alternative way is to adopt some training-free approaches like majority voting \citep{wang2022self} or debate-based approaches \citep{du2023improving}, which have shown effectiveness in aggregating opinions and resolving conflicts between multiple reasoning trajectories. 

Nevertheless, we found that both methods show limitations when applied to mathematical process error identification tasks. Majority voting often fails when errors are identified by only a minority of LLMs \citep{huang2024mirror}. Debate-based approaches sometimes struggle due to an asymmetry in mathematical reasoning: erroneous reasoning paths tend to generate lengthy, seemingly logical justifications, while correct reasoning paths provide only simple justifications. This asymmetry can cause debate methods to favor incorrect justifications, as more elaborate (though flawed) arguments may overshadow simple (but correct) justifications.

To address these limitations, we develop a simple but effective training-free approach to enhance process error identification capabilities. The intuition is to leverage the consistency between a sequence of self-reflection actions because the LLMs should be more likely to remain consistent and confident when asked to review correct validations. As shown in Figure \ref{fig:1}, we propose the \textbf{Temporal Consistency} method, where each LLM iteratively checks its identifications, and the final output is only produced when multiple LLMs demonstrate consistent self-checking over time, effectively reducing unstable incorrect identifications.

\begin{figure}[h]
  \centering
  \includegraphics[width=0.95\linewidth]{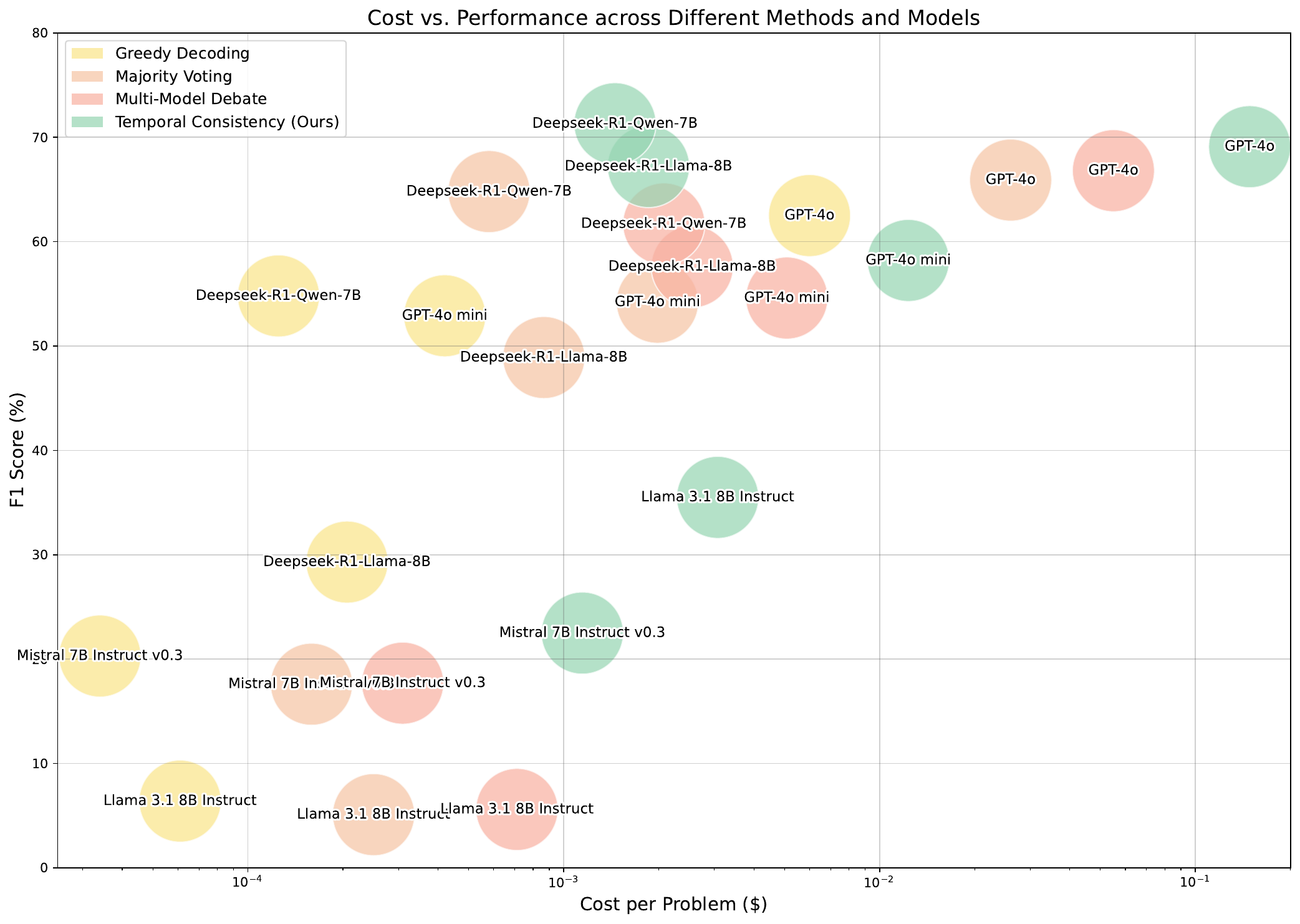}
  \caption{Cost v.s. Performance across different methods and models on ProcessBench. The x-axis (logarithmic scale) shows the cost per problem in dollars (based on OpenRouter pricing \protect\footnotemark), while the y-axis shows the F1 Score percentage.}
  \label{fig:cost_performance}
\end{figure}

We further evaluate our approach across three annotated mathematical step datasets, PRM800K \citep{lightman2023let}, ProcessBench \citep{zheng2024processbenchidentifyingprocesserrors}, and MathCheck$^{*}$\footnote{We use MathCheck$^{*}$ to denote a balanced dataset that combines MathCheck's process judging problems (containing only incorrect solutions) with problems with correct solutions from ProcessBench.} \citep{zhou2024your}. Our experiments demonstrated consistent performance gains across different models, benchmarks, and difficulty levels. We then conducted experiments on R1 distilled models \citep{guo2025deepseek}, where our method achieved remarkable improvements: as shown in Figure \ref{fig:0} for Deepseek-R1-Distill-Llama-8B, improvements of \textbf{46.6\%} on MathCheck$^*$, \textbf{37.9\%} on ProcessBench, and \textbf{29.0\%} on PRM800K; for Deepseek-R1-Distill-Qwen-7B, improvements of \textbf{3.5\%} on MathCheck$^*$, \textbf{16.5\%} on ProcessBench, and \textbf{11.5\%} on PRM800K; for Deepseek-R1-Distill-Qwen-14B, improvements of \textbf{3.7\%} on MathCheck$^*$, \textbf{10.6\%} on ProcessBench, and \textbf{8.6\%} on PRM800K. Notably, our method enables distilled 7B/8B models to achieve \textbf{71.3\%/67.2\%} on ProcessBench, surpassing all existing 70B/72B models and GPT-4o reported in \citet{zheng2024processbenchidentifyingprocesserrors}. With our method applied, the distilled 14B model demonstrates performance comparable to Deepseek-R1's. As shown in Figure \ref{fig:cost_performance}, our Temporal Consistency method establishes a new type of test-time scaling law. Unlike conventional approaches that scale by increasing the number of parallel samples, our method scales through iterative refinement over time (temporal dimension). 

\section{Methodology}\label{algorithm}

In this section, we introduce our method that utilizes multiple rounds of validation to improve identification accuracy. We begin by defining the process error identification task. 
\paragraph{Task Definition}
Given a problem \(P\) and its step-by-step solution \(S = \{s_0, s_1, \dots, s_{n-1}\}\), where each \(s_i\) represents the \(i\)-th solution step, our objective is to identify the first incorrect step, if any, and output a location index \(\texttt{loc} \in \{-1, 0, \dots, n-1\}\). Here, \(\texttt{loc} = -1\) indicating that all steps are correct, while for $\texttt{loc}\geq 0,$ \(s_{\texttt{loc}}\) represents the first incorrect step.

We now introduce the \textbf{Temporal Consistency} algorithm. This method adds a temporal dimension to the verification process by having each LLM consider its own previous assessment, leveraging consistency in a sequence of self-reflection. We employ \(K\) LLMs as verifiers, denoted by \(\mathrm{LLM}^1, \dots, \mathrm{LLM}^K\). The algorithm has three phases:
\footnotetext{https://openrouter.ai}
\begin{figure*}[t]
\centering
\includegraphics[width=1\textwidth]{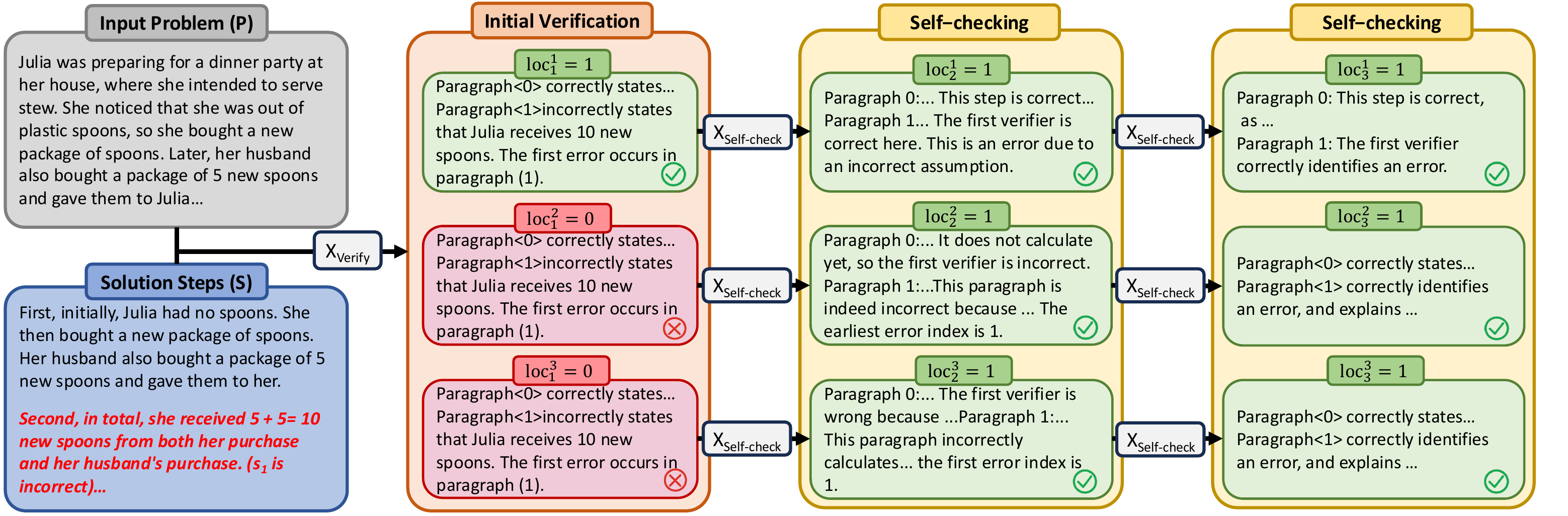}
\caption{Example of the self-checking process: The first error occurred in step 1. Initially, two LLMs incorrectly identified the first incorrect step, while one correctly located the first incorrect step. 
After self-checking, all LLMs achieve the correct identification.
}
\label{fig:2}
\end{figure*}
\paragraph{Initial Verification Phase} 
For each \(i \in \{1, \dots, K\}\), given a problem \(P\), a solution \(S\), and a designated process error identification prompt \(X_{\text{Verify}}\), LLM\(^{i}\) examines the solution step by step. It identifies the location of the first incorrect step \(\texttt{loc}^i_1\), and provides the corresponding reasoning response \(\texttt{res}^i_1\):
\[
(\texttt{loc}^i_1, \texttt{res}^i_1) = \mathrm{LLM}^i(P, S, X_{\text{Verify}})
\]
These initial verifications establish a set of independent assessments.
\paragraph{Iterative Self-checking Phase} 
For time steps \(t \geq 2\), let \((\texttt{loc}^i_{t-1}, \texttt{res}^i_{t-1})\) represent the verification results from the previous iteration for each \(i \in \{1, \dots, K\}\). With a designated self-verification prompt \(X_{\text{Self-check}}\), LLM\(^{i}\) performs a subsequent self-assessment:
\begin{align*}
    (\texttt{loc}^i_t, \texttt{res}^i_t) = \mathrm{LLM}^i(P, S, X_{\text{Self-check}}, \texttt{loc}^i_{t-1}, \texttt{res}^i_{t-1}).
\end{align*}
The distinction between the initial verification phase and the self-checking phase is incorporating previous verification results to provide additional context. This temporal dependency enables the LLMs to potentially correct initial misidentifications. Figure \ref{fig:2} illustrates the self-checking mechanism.

\paragraph{Convergence Check} 
After each iteration $t$, the algorithm determines the majority identification \(\overline{\texttt{loc}}_t\) by applying a majority voting function:
\begin{align}\label{eq:mv}
    \text{MajorityVote}(\texttt{loc}_t^1,\ldots,\texttt{loc}_t^K)=\operatorname*{argmax}_{\texttt{loc} \in \{-1,\cdots,n-1\}}\left|\{i : \texttt{loc}^i_t = \texttt{loc}\}\right|.
\end{align}
This function aggregates the verification outcomes from \(K\) different LLMs and returns the error step that is most frequently identified. Specifically, \(\left|\{i : \texttt{loc}^i = \texttt{loc}\}\right|\) counts the number of LLMs that have identified step \(\texttt{loc}\) as incorrect. The algorithm then evaluates the stability of these identifications across all LLMs. Let \(p_t\) be the proportion of agents supporting \(\overline{\texttt{loc}}_t\), formally defined as
\begin{align}\label{eq:portion}
    p_t = \frac{\left|\{ i : \texttt{loc}^i_t = \overline{\texttt{loc}}_t \}\right|}{K}.
\end{align}
When sufficient stability and consensus are reached, the algorithm terminates and outputs the final identification. 
Detailed stopping conditions defined with $\texttt{loc}_t$ and $p_t$ are provided in Section \ref{sec:stop}.

This approach leverages the strengths of multiple independent verifications and consistency across the temporal dimension. By allowing each LLM to build on its previous assessments while remaining isolated from others, the algorithm minimizes the risk of reinforcing arguments that appear plausible but are incorrect. The complete algorithm is detailed in Algorithm \ref{alg:self_verification}.


\subsection{Stopping Criteria}\label{sec:stop}
In practice, most agents converge to an identification within just a few rounds, making further self-checks computationally redundant. To enhance efficiency, we propose a heuristic stopping criterion that permits early termination for "high confidence" problems while allowing continued self-checking for "low confidence" problems.
\begin{algorithm}[t]
\caption{Temporal Consistency}
\label{alg:self_verification}
\begin{algorithmic}
    \STATE {\bfseries Input:} Problem $P$, solution $S$, number of LLMs $K$, initial verification prompt $X_{\text{Verify}}$, self-checking prompt $X_{\text{Self-check}}$, consistency requirement $q$, max rounds $T$.   
    \STATE {\bfseries /* Initial Verification Phase */}
    \FOR{$i=1$ {\bfseries to} $K$ in parallel}
        \STATE $(\texttt{loc}^i_1, \texttt{res}^i_1) \gets \mathrm{LLM}^i(P,S, X_{\text{Verify}})$ 
    \ENDFOR
    
    \STATE {\bfseries /* Iterative Self-checking Phase */}
    \FOR{round $t=2$ {\bfseries to} $T$}
        \FOR{LLM $i=1$ {\bfseries to} $K$ in parallel}
            \STATE $(\texttt{loc}^i_t, \texttt{res}^i_t) \gets \mathrm{LLM}^i(P,S,X_{\text{Self-check}}, \texttt{loc}^i_{t-1}, \texttt{res}^i_{t-1})$
        \ENDFOR
        \STATE $\overline{\texttt{loc}}_t \gets \text{MajorityVote}(\texttt{loc}^1_t, ..., \texttt{loc}^K_t)$
        \STATE $p_t \gets |\{i: \texttt{loc}^i_t = \overline{\texttt{loc}}_t\}| / K$ 
        
        \IF{$t \geq q$}
            \STATE $\text{stable} \gets \bigwedge_{j=0}^{q-2}(\overline{\texttt{loc}}_{t-j} = \overline{\texttt{loc}}_{t-q+1}) $
            \STATE $\text{growing} \gets \bigwedge_{j=0}^{q-2}(p_{t-j} \geq p_{t-j-1})$
            \IF{$\text{stable}$ {\bfseries and} $\text{growing}$}
                \STATE {\bfseries return} $\overline{\texttt{loc}}_t$
            \ENDIF
        \ENDIF
    \ENDFOR
    \STATE {\bfseries return} $\overline{\texttt{loc}}_T$ 
    \COMMENT{Return final majority if max rounds reached}
\end{algorithmic}
\end{algorithm}

For any round \(t \in \{1,\ldots,T\}\), let \(\overline{\texttt{loc}}_t\) denote the majority identification defined in \eqref{eq:mv}, and $p_t$ be the proportion of agents supporting \(\overline{\texttt{loc}}_t\) defined in \eqref{eq:portion}. Based on these definitions, we design two stopping conditions over \(q\) consecutive rounds, where $q$ is a given consistency requirement:

1. \textbf{Majority Stability}:
   \begin{align*}
       &\overline{\texttt{loc}}_{t-q+1} = \overline{\texttt{loc}}_{t-q+2}=\cdots  = \overline{\texttt{loc}}_t,
   \end{align*}
   
2. \textbf{Growing Consensus}:
   \[
   p_{t-q+1}\leq p_{t-q+2} \leq \cdots \leq p_t.
   \]

The majority stability condition requires that the majority identification remains unchanged over the past \(q\) rounds, ensuring a consistent outcome in majority voting. Concurrently, the growing consensus condition needs the proportion of agents supporting the majority identification to not decrease across these \(q\) rounds. The underlying intuition is that the correct answer should be identified with "increasing confidence" over the past \(q\) rounds.

The algorithm terminates when both conditions are satisfied or when the maximum number of rounds \(T\) is reached. The consistency requirement \(q\) is a parameter that can be adjusted according to task-specific requirements.
\begin{figure*}[t]
    \centering
    \includegraphics[width=1.0\linewidth]{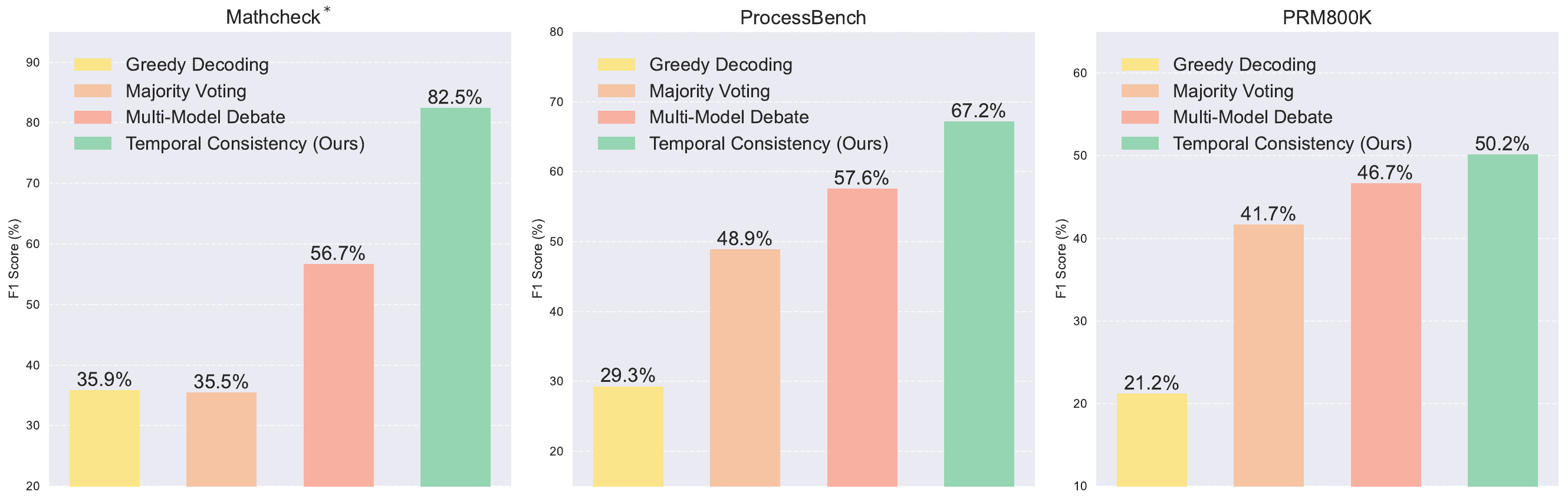}
    \caption{Performance comparison across three datasets (Mathcheck$^*$, ProcessBench, and PRM800K). Our Temporal Consistency approach (green) consistently outperforms baseline methods, including greedy decoding (yellow), majority voting (orange), and multi-model debate (red).}
    \label{fig:3}
\end{figure*}
\subsection{Comparison with Existing Methods}
Existing majority voting approaches \citep{cobbe2021training, li2022competition, wang2022self} perform multiple generations simultaneously, essentially scaling horizontally to enhance stability. In contrast, our method allows each LLM to build upon its previous assessments, achieving vertical scaling over time. This sequential self-reflection enables each verification to benefit from prior insights.

Moreover, our approach differs from multi-model debate methods \citep{du2023improving} in treating LLM independence. Although debate methods allow models to exchange information, thus enabling them to see other agents' answers and gain additional perspectives, this openness risks influence from persuasive yet incorrect arguments. For further illustration, an example can be found in Appendix \ref{app:ex}. In contrast, our method maintains strict isolation between LLMs. Each LLM focuses solely on its own reasoning process, thereby reducing the risk of propagating elaborate but erroneous arguments.
%
\section{Experiments}
\label{sec:experiments}
\subsection{Experimental Setup}
\paragraph{Dataset} We evaluate our method on ProcessBench \citep{zheng2024processbenchidentifyingprocesserrors}, a comprehensive dataset combining multiple mathematical problem-solving benchmarks. The dataset consists of 3,400 problems from four sources: 400 from GSM8K \citep{cobbe2021training}, 1,000 from MATH dataset \citep{hendrycks2021measuring}, 1,000 from OlympiadBench \citep{he2024olympiadbench}, and 1,000 from Omni-MATH \citep{gao2024omni}. Each problem includes both generated solutions and human-annotated processes. Additionally, we incorporate 516 process judging problems based on GSM8K from MathCheck \citep{zhou2024your} and 300 randomly selected problems based on MATH dataset from PRM800K \citep{lightman2023let}. Since the process judging problem in MathCheck only contains incorrect solutions, we combine it with the GSM8K problems with correct steps from ProcessBench to create a balanced dataset, which we denote as MathCheck$^*$. For PRM800K, we consider both 0 and 1 annotations as correct steps and -1 as incorrect steps. We evaluate the F1 score for all benchmarks, which is the harmonic mean of the accuracies on incorrect and correct samples.
\begin{table*}[t]
\centering
\resizebox{\textwidth}{!}{%
\begin{tabular}{llcccc}
\hline
Model & Method & Mathcheck$^*$ & ProcessBench & PRM800K \\
\hline
\multirow{4}{*}{GPT-4o mini} 
& Greedy Decoding & 78.8 & 52.9 & 34.0 \\
& Majority Voting & 80.4 & 54.2 & 37.9  \\
& Multi-Model Debate & 79.9 & 54.6 & 38.0 \\
\rowcolor{lightblue}& Temporal Consistency (Ours) & \textbf{84.8} & \textbf{58.2}  & \textbf{39.0}  \\
\hline
\multirow{4}{*}{GPT-4o}
& Greedy Decoding & 87.3 & 62.5 & 41.6  \\
& Majority Voting & 89.0 & 65.9 & 42.6  \\
& Multi-Model Debate & 90.8 & 66.8 & 50.7  \\
\rowcolor{lightblue}& Temporal Consistency (Ours) & \textbf{91.8} & \textbf{69.1} & \textbf{51.6}  \\
\hline
\multirow{4}{*}{Llama 3.1 8B Instruct}
& Greedy Decoding & 13.3 & 6.4 & 2.4 \\
& Majority Voting & 5.9 & 5.1 & 6.8  \\
& Multi-Model Debate & 6.8 & 5.6 & 2.6  \\
\rowcolor{lightblue}& Temporal Consistency (Ours) & \textbf{60.2} & \textbf{35.5} & \textbf{22.1} \\
\hline
\multirow{4}{*}{Mistral 7B Instruct v0.3}
& Greedy Decoding & 26.4 & 20.3 & 13.0  \\
& Majority Voting & 26.3 & 17.6 & 12.1  \\
& Multi-Model Debate & 26.2 & 17.7 & 12.1  \\
\rowcolor{lightblue}& Temporal Consistency (Ours) & \textbf{37.4} & \textbf{22.5} & \textbf{13.3} \\
\hline
\end{tabular}%
}
\caption{Performance comparison across different models. Numbers represent F1 score (\%). The best performance for each model is highlighted in bold. Our method consistently outperforms baselines across all models and benchmarks.}
\label{tab:cross_model}
\end{table*}
\begin{table*}[h]
\centering
\resizebox{\textwidth}{!}{%
\begin{tabular}{llcccc}
\hline
Model & Method & Mathcheck$^*$ & ProcessBench & PRM800K \\
\hline
\multirow{4}{*}{Deepseek-R1-Qwen-7B}
& Greedy Decoding & 86.0 & 54.8 & 46.2  \\
& Majority Voting & 89.3 & 64.8 & 55.1 \\
& Multi-Model Debate & 84.8 & 61.7 & 51.2  \\
\rowcolor{lightblue}& Temporal Consistency (Ours) & \textbf{89.5} & \textbf{71.3} & \textbf{57.7} \\
\hline
\multirow{4}{*}{Deepseek-R1-Llama-8B}
& Greedy Decoding & 35.9 & 29.3 & 21.2  \\
& Majority Voting & 35.5 & 48.9 & 41.7  \\
& Multi-Model Debate & 56.7 & 57.6 & 46.7  \\
\rowcolor{lightblue}& Temporal Consistency (Ours) & \textbf{82.5} & \textbf{67.2} & \textbf{50.2} \\
\hline
\end{tabular}%
}
\caption{Performance comparison of Deepseek R1 distilled models on three benchmarks. Numbers represent F1 score (\%). The best performance for each model is highlighted in bold.}
\label{tab:distill}
\end{table*}


\paragraph{Baseline Methods} We compare our approach against three baseline methods:
(1) Verification with greedy decoding \citep{zhang2022automatic}, where a single agent generates a verification deterministically,
(2) Majority voting among multiple agents \citep{wang2022self}, where multiple agents independently generate verifications, and the final decision is made based on majority voting and
(3) Verification with debate-based reasoning \citep{du2023improving}, where multiple agents generate verifications independently, and they will receive the answer from the other agents and then generate a new identification. 
\paragraph{Parameter Setting} To ensure a fair comparison, we employ 5 parallel agents in each of the three methods: majority voting, debate-based verification, and our Temporal Consistency approach. Following \citet{du2023improving}, the debate method proceeds in two rounds: an initial verification round followed by a debate round. Our method implements convergence criteria requiring stability across 3 consecutive rounds, with a maximum of 10 rounds. We use Deepseek-R1-Llama-8B \citep{guo2025deepseek} in all our experiments except those in Table \ref{tab:cross_model}, Table \ref{tab:distill} and Figure \ref{fig:0}. Appendix \ref{sec:detail} shows complete experimental configurations and implementation details.
\subsection{Main Results}

\paragraph{Improvement over Diverse Dataset}
Figure \ref{fig:3} presents the performance comparison across three datasets for Deepseek-R1-Llama-8B. Our Temporal Consistency approach consistently outperforms baseline methods across all evaluation settings.

On Mathcheck$^*$, our method achieves an F1 score of 82.5\%, showing an improvement of 46.6\% over greedy decoding and 25.8\% over multi-model debate. For ProcessBench, we observe consistent improvements with our method achieving 67.2\% F1 score, compared to 29.3\% for greedy decoding and 57.6\% for multi-model debate. On PRM800K, our method maintains its advantage with 50.2\% F1 score, showing a 29.0\% improvement over greedy decoding.

\paragraph{Improvement over Different Base Models} To demonstrate the generalizability of our approach, we conducted experiments across different language models, including GPT-4o mini, GPT-4o \citep{hurst2024gpt},  Llama 3.1 8B Instruct \citep{grattafiori2024llama} and Mistral 7B Instruct \citep{jiang2023mistral}. We evaluated these models on Mathcheck$^*$, ProcessBench, and PRM800K. As shown in Table \ref{tab:cross_model}, our Temporal Consistency method consistently outperforms baseline methods across all 
 benchmarks. This consistent performance across different models demonstrates the effectiveness of our approach.
\paragraph{Improvement for Distilled Models} We further evaluate our method and the baseline methods on the recently released Deepseek R1 distilled models \citep{guo2025deepseek}, including DeepSeek-R1-Distill-Qwen-7B and DeepSeek-R1-Distill-Llama-8B. As shown in Table \ref{tab:distill}, our Temporal Consistency method demonstrates remarkable effectiveness on 7B/8B-scale models, achieving 71.3\% and 67.2\% accuracy on ProcessBench with DeepSeek-R1-Distill-Qwen-7B and DeepSeek-R1-Distill-Llama-8B respectively, surpassing GPT-4o (69.1\%) and all 70B/72B models reported in \citet{zheng2024processbenchidentifyingprocesserrors}, including Llama-3.3-70B-Instruct (58.0\%), Qwen2.5-Math-72B-Instruct (45.5\%) and Qwen2.5-72B-Instruct (61.2\%) \citep{yang2024qwen2}. 
\subsection{Additional Analysis}

\begin{figure}[t]
  \centering
  \begin{minipage}[t]{0.48\linewidth}
     \centering
    \includegraphics[width=\linewidth]{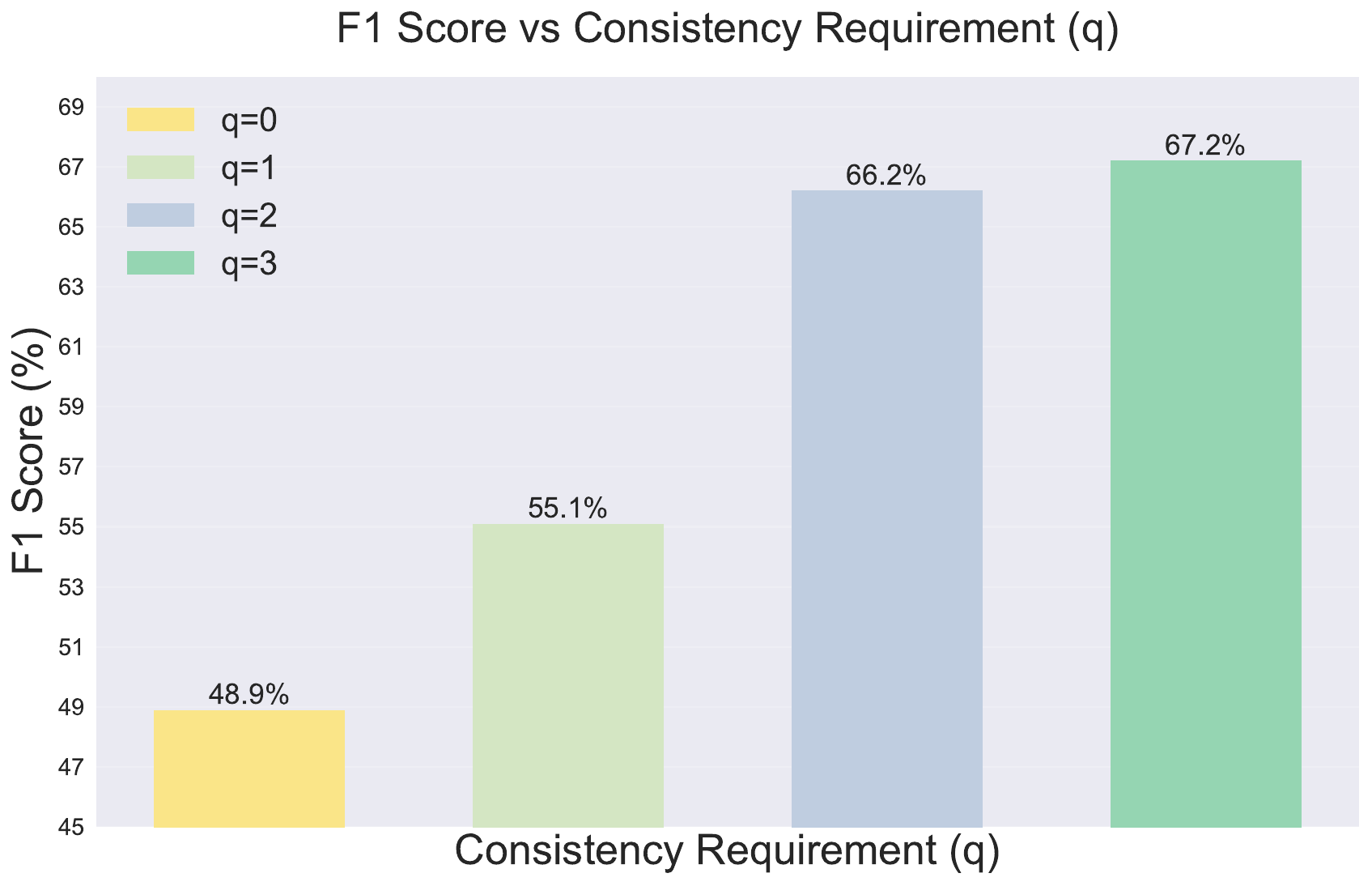}
    \caption{Performance comparison across different consistency requirements on ProcessBench for Deepseek-R1-Llama-8B. Higher consistency requirements, indicating stricter stability requirements, correlate with improved F1 scores.}
    \label{fig:stopping_param}
  \end{minipage}
  \hfill
  \begin{minipage}[t]{0.48\linewidth}
    \centering
    \includegraphics[width=\linewidth]{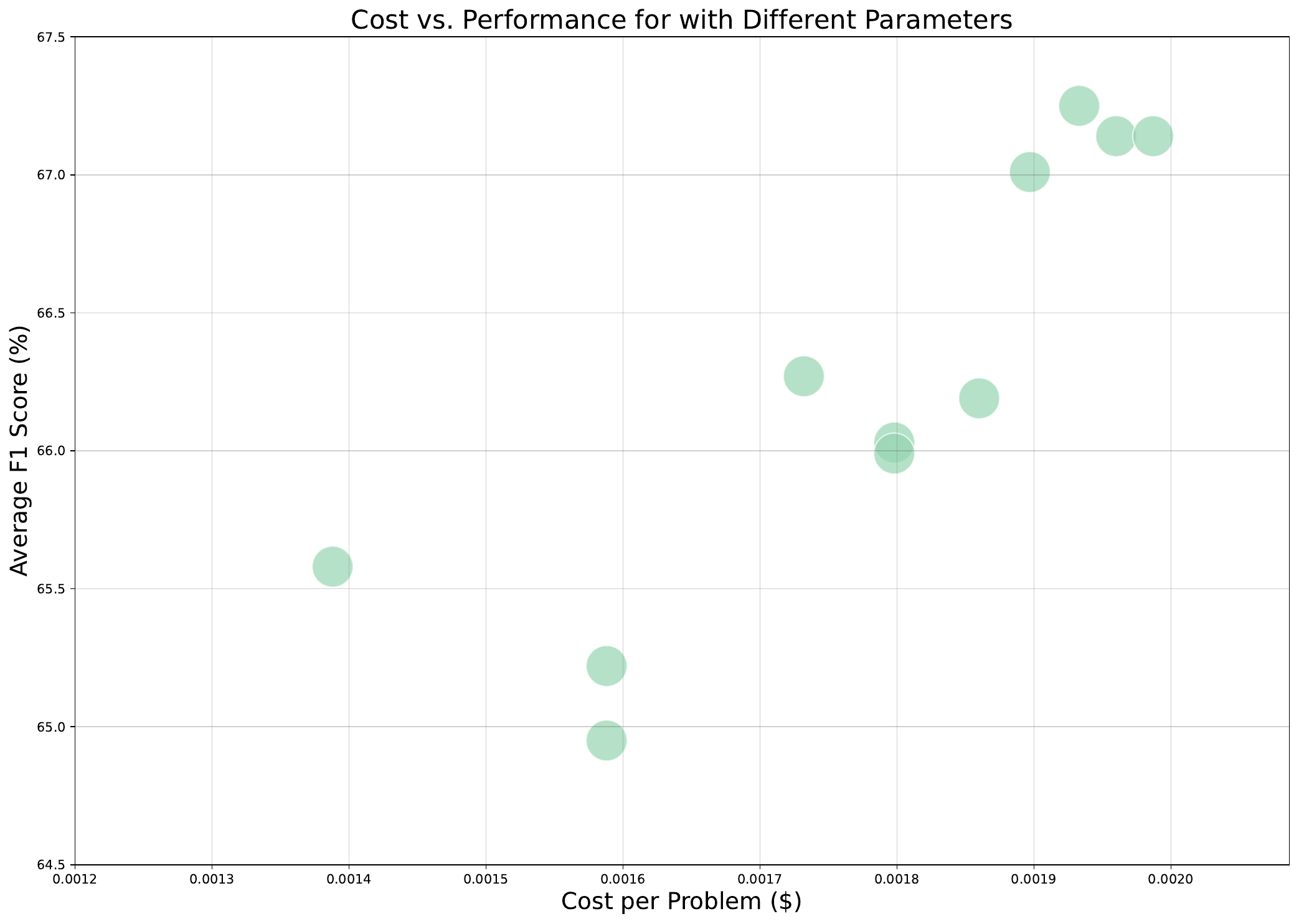}
    \caption{Cost-performance analysis of our method with different parameter configurations (max rounds and consistency requirement) on ProcessBench for Deepseek-R1-Llama-8B. The horizontal axis shows the cost per problem, while the vertical axis shows the average F1 score. As the computational budget increases, we observe improved performance, demonstrating the effectiveness of additional test-time scaling computation resources.}
    \label{fig:con}
  \end{minipage}
\end{figure}
\paragraph{Different Choice of Consistency Requirement} We investigated the impact of different consistency requirements on model performance using ProcessBench. As shown in Figure \ref{fig:stopping_param}, we experimented with consistency requirements ranging from $0$ to $3$, where higher values indicate stricter requirements for output stability. The F1 score demonstrates a consistent upward trend as the consistency requirement increases, starting from $48.9$\% without the self-checking requirement (parameter = $0$) and reaching $67.2$\% with the strictest stability requirement (parameter = $3$). This correlation suggests that requiring more stable outputs through multiple verification rounds leads to more accurate results.

\begin{figure}[t]
  \centering
  \begin{minipage}[t]{0.48\linewidth}
    \centering
    \includegraphics[width=\linewidth]{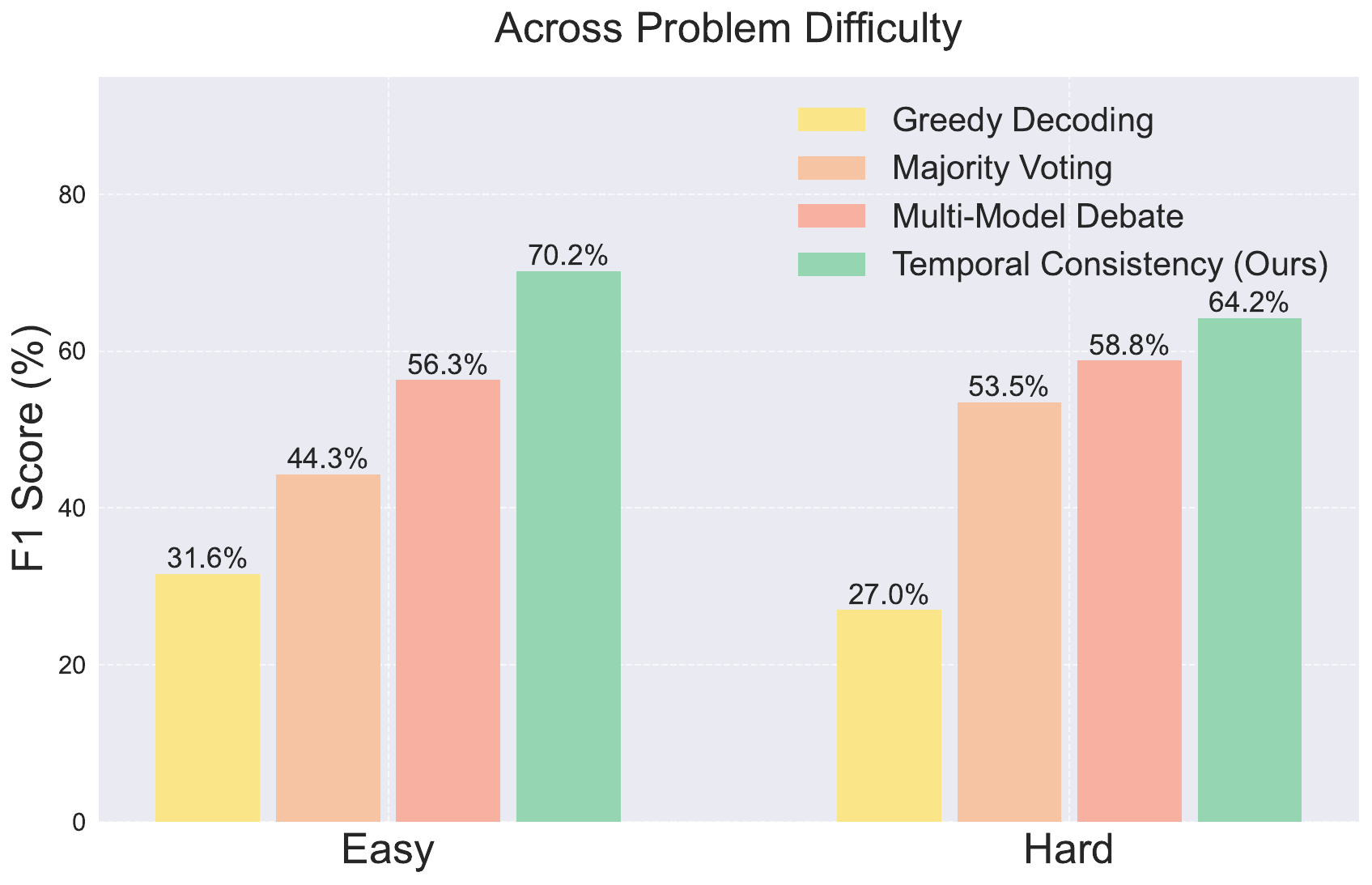}
    \caption{Performance comparison across problem difficulty levels. Problems are categorized as Easy (from GSM8K and MATH) or Hard (from OlympiadBench and Omni-MATH). Our method shows particular advantages on harder problems, maintaining more stable performance than baseline approaches.}
    \label{fig:difficulty_comparison}
  \end{minipage}
  \hfill
  \begin{minipage}[t]{0.48\linewidth}
    \centering
    \includegraphics[width=\linewidth]{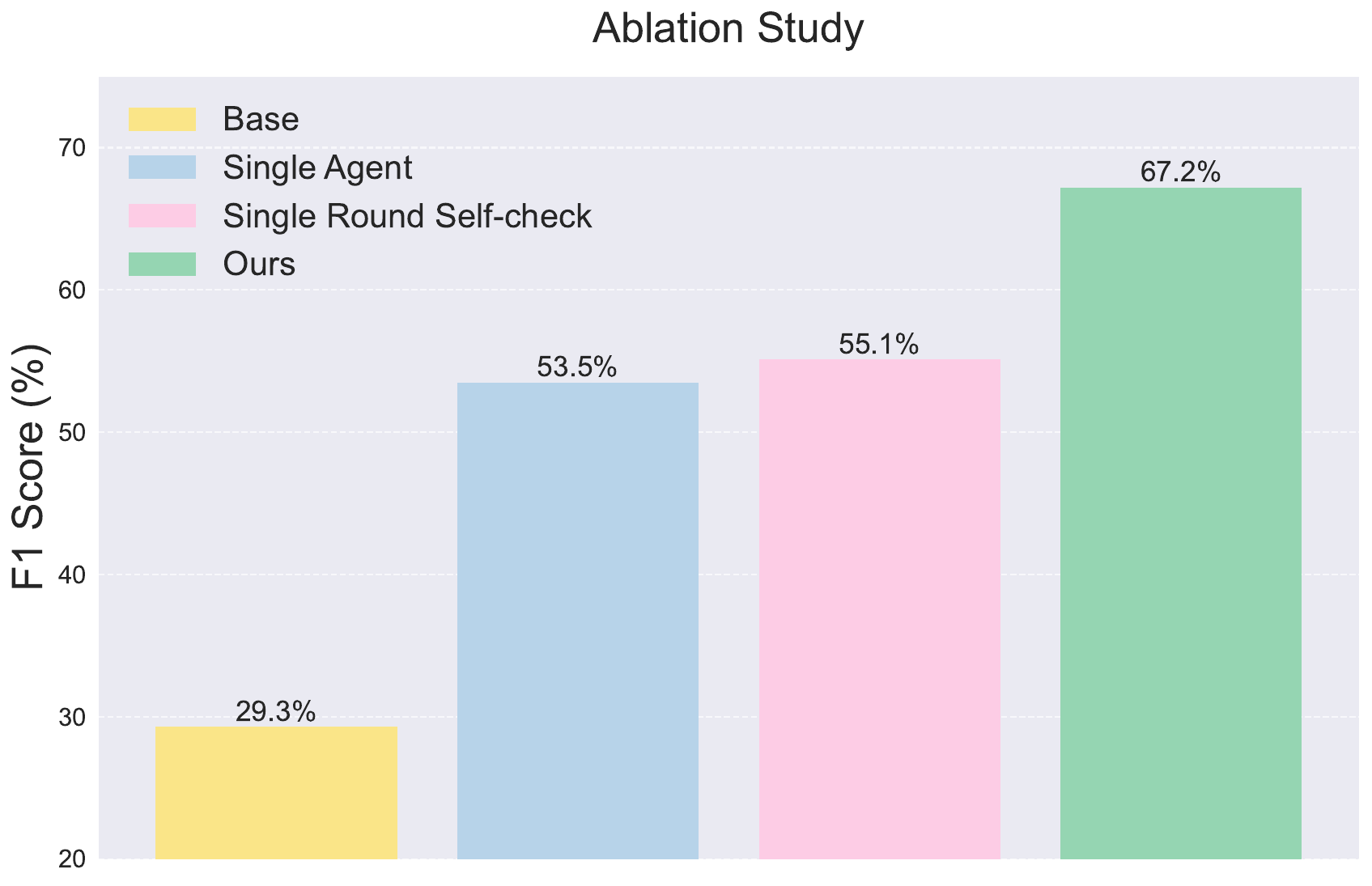}
    \caption{Ablation study results for ProcessBench demonstrating the effectiveness of both iterative generation and multi-agent components, with their combination yielding the best performance.}
    \label{fig:ablation}
  \end{minipage}
\end{figure}

\paragraph{Performance Across Problem Difficulty}
To analyze our method's effectiveness across varying complexity levels, we categorized ProcessBench problems into two groups following the difficulty definition in \citet{zheng2024processbenchidentifyingprocesserrors}: Easy (derived from GSM8K and MATH) and Hard (derived from OlympiadBench and Omni-MATH). Figure \ref{fig:difficulty_comparison} illustrates the performance comparison across these categories.

All methods demonstrate strong performance on easy problems, with our approach achieving 70.2

\paragraph{Cost-Performance Analysis} To understand the trade-offs between computational resources and verification performance, we conducted experiments with various parameter configurations of our method. Figure \ref{fig:con} illustrates how performance scales with increased computational budget across different parameter settings. We observe a general trend where higher computational investment yields better verification results.


\paragraph{Ablation Study} To understand the contribution of each component in our approach, we conducted an ablation study on ProcessBench, with results shown in Figure \ref{fig:ablation}. We evaluated four configurations: the greedy decoding method, Temporal Consistency without multi-agent, self-checking without iterative generation, and our method. The results demonstrate that both components contribute to the overall performance. Starting from the base F1 score of 29.3\%, each component independently improves performance, with the multi-agent self-checking and iterative mechanisms contributing improvements of 24.2\% and 25.8\%, respectively. The combination achieves the best performance with an F1 score of 67.2\%. 
\section{Related Work}\label{sec:related}
\paragraph{Datasets and Benchmarks for Process Error Detection} Process error detection in mathematical reasoning requires annotations at the step level, currently available in three major datasets. PRM800K \citep{lightman2023let} pioneered this direction by providing human-annotated reasoning steps based on the MATH dataset \citep{hendrycks2021measuring}, focusing on high school and college-level mathematics. MathCheck \citep{zhou2024your} extends this approach to elementary mathematics by synthesizing solutions with incorrect steps from GSM8K problems \citep{cobbe2021training}, offering a systematic evaluation of step-by-step verification. Most recently, ProcessBench \citep{zheng2024processbenchidentifyingprocesserrors} expands the coverage of mathematical difficulty by providing expert-annotated solution steps across four distinct datasets: GSM8K, MATH, and notably, OlympiadBench \citep{he2024olympiadbench} and Omni-MATH \citep{gao2024omni} for competition and olympiad-level challenges. Our experimental evaluation across these benchmarks provides comprehensive insights into our method's effectiveness from basic arithmetic to advanced mathematical reasoning.

\paragraph{Process Error Identification Methods} Approaches to error detection in language models can be categorized into two main streams. The first focuses on training specialized verification models, such as process reward models \citep{lightman2023let, luo2024improve, setlur2024rewarding, wang2024math, zhang2024generative} and finetuned language models \citep{cobbe2021training, kang2024mindstar,  zheng2024critic, yang2024supercorrect, tang2025enabling, luo2025ursa, guan2025rstar}. While these training-based methods have shown promising results, they require additional training data and significant computational resources, especially for larger models. The second stream explores inference-time verification through prompting techniques like self-reflection \citep{miao2023selfcheckusingllmszeroshot, madaan2024self}. Recent work has demonstrated that language models often struggle to correct errors without external feedback \citep{huang2023large, kamoi2024can}. Similar to self-reflection work \citep{madaan2024self,yang2024supercorrect}, which iteratively generates improvement suggestions, our method employs an iterative process. 

Rather than training new models, we focus on utilizing existing models more effectively. However, our Self-check method can also be applied to trained verification models to improve their accuracy potentially.

\paragraph{More General Reasoning Methods} 
The broader field of reasoning in language models has explored various frameworks to enhance problem-solving capabilities and solution reliability. Chain-of-Thought prompting \citep{wang2022self} and its variants like Tree-of-Thought \citep{yao2024tree} and Buffer-of-Thought \citep{yang2024buffer} have demonstrated that explicitly articulating intermediate reasoning steps improves model performance on complex reasoning tasks, and \citet{zhang2024generative} further validates the effectiveness of reasoning in verification tasks. Predesigned reasoning structures \citep{zhang2022automatic, besta2024graph, yang2024supercorrect} have also shown promise in improving mathematical capabilities by guiding LLMs to think along predefined trajectories. Multi-agent approaches such as debate mechanisms \citep{du2023improving, subramaniam2025multiagent} enable models to critically examine solutions through structured discussions, while majority voting methods \citep{wang2022self} generate multiple independent solutions and aggregate them through majority voting to enhance reliability. While each approach offers unique advantages, they demonstrate the importance of structured reasoning processes in improving model performance.
\paragraph{Test Time Scaling} Recent studies have demonstrated that leveraging multiple samples during inference can significantly enhance model performance \citep{hurst2024gpt, guo2025deepseek,yang2025reasonflux}. Through iterative refinement, models incorporate feedback from previous generations to guide subsequent outputs \citep{snell2024scalingllmtesttimecompute, hou2025advancing, lee2025evolving}. While early approaches focused on simple majority voting strategies \citep{wang2022self}, subsequent research has advanced towards more sophisticated techniques, particularly in search-based methods \citep{khanov2024args, wan2024alphazero,yang2025reasonflux}. The field has evolved with hybrid frameworks that seamlessly integrate tree-based search with sequential approaches \citep{wu2024inference, snell2024scalingllmtesttimecompute, qiu2024treebon, gandhi2024stream}. \citet{liu2025can} conducted a study on optimizing test-time computation scaling across various policy models and problem complexities. Most closely related to our approach, \citet{muennighoff2025s1} achieved substantial improvements in competition math questions by implementing parallel self-reflection on historical interactions.


\section{Conclusion}
We presented an Temporal Consistency approach for improving mathematical process error identification in language models. Our method leverages temporal consistency patterns in verification behavior, allowing LLMs to recheck their judgments through multiple rounds. We demonstrated how this approach effectively improves verification accuracy across different models and problem types through empirical evaluation.

Our key insight is that the temporal stability of verifications can serve as a reliable indicator of correctness. This finding opens new directions for developing methods focusing on consistency over time rather than agreement across agents. Our results suggest that incorporating temporal dynamics can enhance the reliability of mathematical reasoning methods.

\section*{Limitations}\label{sec:lim}
While our Temporal Consistency approach demonstrates consistent improvements across different settings, it has several limitations. First, the method requires multiple verification rounds, leading to increased computational costs. Second, our method has been evaluated in the context of mathematical tasks, and it may not hold true for other reasoning tasks. 
\bibliographystyle{unsrtnat}
\bibliography{ref}

\newpage
\appendix
\newpage

\section{Implementation Details}\label{sec:detail}

We use the gpt-4o-2024-08-06 API for GPT-4o and gpt-4o-mini API for GPT-4o-mini. We use Together API for the Deepseek-R1 model. All experiments can be performed on a single NVIDIA H100 GPU. 

In the first round of all methods, the generation process was conducted using a temperature setting 0.7. The subsequent rounds vary slightly between closed-source and open-source models, with the following specifics:

\begin{itemize}
    \item \textbf{Closed-source models:} For the debate method and our approach in later rounds, the temperature was set to 1.

    \item \textbf{Open-source models:} We used a fixed random seed of 42 throughout the experiments. For the debate method and subsequent rounds of our approach, the temperature was set to 0.7, and top-p=0.8, top-k=40. 
\end{itemize}

\subsection{Prompting Strategy for Initial Verification}

In the first round of all methods, we utilized the verification prompts provided in \citet{zheng2024processbenchidentifyingprocesserrors}. The prompt format for the initial generation was:

\begin{quote}
\texttt{The following is a math problem and a solution (split into paragraphs, enclosed with tags, and indexed from 0):}

\texttt{[Math Problem]}

\texttt{\{problem\}}

\texttt{[Solution]}

\texttt{\{tagged\_response\}}

\texttt{Your task is to review and critique the solution paragraph by paragraph. Once you identify an error in a paragraph, return the index where the earliest error occurs. Otherwise, return the index of -1 (which typically denotes "not found").}

\texttt{Please put your final answer (i.e., the index) in \textbackslash boxed\{\}.}
\end{quote}

\subsection{Debate Method Prompt Adaptation}

The debate method is not designed for the verification task. To adapt it to our context, we combined the prompts for initial verification with those described in the appendix of \citet{du2023improving}. The adapted prompt is as follows:

\begin{quote}
\texttt{These are the solutions to the problem from other agents:} \\
\texttt{One agent solution: \texttt{\{res['reason']\}}}

\texttt{Using the solutions from other agents as additional information, please analyze this solution and end with the earliest error index in \textbackslash boxed\{\}, or -1 in \textbackslash boxed\{\} if no errors.}
\end{quote}

\subsection{Prompt for Secondary Verification}

For the subsequent verification rounds in our approach, we designed a custom prompt to ensure thorough checking of the initial LLM’s results. The prompt used is:
















\begin{Verbatim}[breaklines=true]
You are a secondary LLM for math problem solutions. The first verifier's task was to review math solutions paragraph by paragraph and identify the earliest error (if any), returning -1 if no errors were found.

The first verifier may have made mistakes. Your job is to carefully check their work. You will receive:

1. The original math problem

2. The solution steps

3. The first verifier's generated label (paragraph index where they found the first error, or -1)

4. The first verifier's reasoning

IMPORTANT:

1. You must check each paragraph carefully, as if you are re-solving the problem from scratch.

2. If you find any error—no matter how minor—locate the earliest paragraph containing that error.

3. If the solution is correct throughout, only then do you output -1.

4. The first verifier may be wrong. You cannot just accept their result. Always verify carefully and do not hesitate to disagree.

Your output format:

1. Begin with an overview: "Let's check the solution paragraph by paragraph based on the first agent's verification:"

2. For each paragraph (starting from paragraph 0), specify whether it is correct or not based on the first verifier's reason. If an error is found, explain the reason for the error. If correct, explain why it is correct.

3. End with the earliest error index in \boxed, or -1 in \boxed if no errors.
\end{Verbatim}


\section{Examples for Various Methods}\label{app:ex}

\begin{methodbox}{Example for Majority Voting (3 agents)}
\vspace{0.1cm}
\textbf{Problem:} 
\vspace{0.1cm}
Julia was preparing for a dinner party at her house, where she intended to serve stew. She noticed that she was out of plastic spoons, so she bought a new package of spoons. Later, her husband also bought a package of 5 new spoons and gave them to Julia. While Julia was making the stew, she used three of the spoons to sample her stew. Later, when she went to set the table, she had a total of 12 spoons. How many spoons were in the package that Julia bought?

\vspace{0.1cm}
\textbf{Solution Steps:} 
\vspace{0.1cm}

\textbf{Step 0:} 
Let's break down the problem step by step: First, initially, Julia had no spoons. She then bought a new package of spoons. Her husband also bought a package of 5 new spoons and gave them to her.

\textbf{Step 1:} 
Second, in total, she received 5 + 5 = 10 new spoons from both her purchase and her husband's purchase. \textcolor{red}{First Error Step}

\textbf{Step 2:} 
Third, while making the stew, she used 3 of the spoons to sample it, leaving her with 10 - 3 = 7 spoons.

\textbf{Step 3:} 
Fourth, when she went to set the table, she had a total of 12 spoons. We need to find out how many spoons were in the package that Julia bought initially.

\textbf{Step 4:} 
We know she received 10 more spoons after her husband gave her the additional 5 spoons. Since she had 7 spoons left after sampling and ended up with 12 spoons in total, the number of spoons she started with must have been 12 - 7 = 5.

\textbf{Step 5:} 
Therefore, the number of spoons in the package that Julia bought is 5.

\newpage

\textbf{Agent Analysis:}
\vspace{0.1cm}

\textbf{Agent 1:} 
\begin{itemize}
\item Paragraph <0> assumes Julia started with no spoons which isn't stated in problem
\item This is an incorrect assumption as we don't know initial spoon count
\item First error occurs in paragraph <0>
\item Therefore labels error as paragraph 0
\end{itemize}

\vspace{0.1cm}
\textbf{Agent 2:} 
\begin{itemize}
\item Paragraph <0> contains unstated assumption about initial spoon count
\item Making assumptions not given in original problem is an error
\item First error identified in step <0>
\item Labels error as paragraph 0
\end{itemize}

\vspace{0.1cm}
\textbf{Agent 3:}
\begin{itemize}
\item Paragraph <0> correctly states initial conditions
\item Paragraph <1> makes incorrect assumption about total spoons being 10
\item First error found in calculating total spoons without knowing Julia's package
\item Labels error as paragraph 1
\end{itemize}
 
\end{methodbox}

\begin{methodbox}{Example for Change of Opinion After Debate (3 agents)}
\vspace{0.1cm}
\textbf{Problem:} 
\vspace{0.1cm}
Amy is taking a history test. She correctly answers 80\% of the multiple-choice questions, 90\% of the true/false questions, and 60\% of the long-answer questions. The multiple-choice and true/false questions are worth 1 point each, and the long answer questions are worth 5 points each. How many points does Amy score if there are 10 multiple-choice questions, 20 true/false questions, and 5 long answer questions?

\textbf{Solution Steps:} 
\vspace{0.1cm}

\textbf{Step 0:} 
To calculate the total points that Amy scores on her history test, we need to determine how many points she earns from each type of question and then sum those points.

\textbf{Step 1:} 
First, for the multiple-choice questions:
- There are 10 multiple-choice questions, each worth 1 point.
- Amy answers 80\% of them correctly.
- Points from multiple-choice questions = 10 × 0.80 = 8 points.

\textbf{Step 2:} 
Second, for the true/false questions:
- There are 20 true/false questions, each worth 1 point.
- Amy answers 90\% of them correctly.
- Points from true/false questions = 20 × 0.90 = 18 points.

\textbf{Step 3:} 
Third, for the long-answer questions: 
- There are 5 long-answer questions, each worth 5 points.
- Amy answers 60\% of them correctly.
- Points from long-answer questions = 5 × 0.60 × 5 = 15 points.

\textbf{Step 4:} 
Finally, to find the total points:
- Total points = Points from multiple-choice + Points from true/false + Points from long-answer
- Total points = 8 + 18 + 15
- Total points = 41

\textbf{Step 5:} 
Therefore, Amy scores a total of 41 points on her history test.

\textcolor{red}{No Error ($\texttt{loc}=-1$)}
\newpage
\textbf{Agent Analysis:}

\vspace{0.1cm}

\textbf{First Round:}
\begin{itemize}
\item Agent 1: "After reviewing each paragraph step-by-step, no computational errors or mistakes in the logic of the solution are found. Therefore, index -1."
\item Agent 2: "Upon reviewing each paragraph, no errors are found. While the calculation method in paragraph 3 is unconventional, it yields the correct result. Therefore, index -1."
\item Agent 3: "Here, there is an error. The calculation should be 5 × 5 × 0.60 = 15 points, which is correct. However, the error is in the explanation that wrongly states the operation order as 5 × 0.60 × 5. While the final result is correct, the explanation could be misleading. The first minor error in explanation is found in Paragraph <3>. Therefore, index 3."
\end{itemize}

\vspace{0.1cm}
\textbf{Second Round (After Discussion):}
\begin{itemize}
\item Agent 1: "Upon reviewing the solution and considering the feedback from other agents, I maintain that there are no computational errors affecting the final result. Index -1."
\item Agent 2: "After reviewing the solutions from other agents, I now see that while the calculation gives the correct result, the explanation in paragraph 3 can be misleading. As Agent 3 pointed out, the operation order should be more clearly stated. Therefore, I revise my assessment to index 3."
\item Agent 3: "Upon reviewing the solutions provided by other agents, I maintain my position. The calculation in paragraph 3, while giving the correct result, has a misleading explanation that could cause confusion. Index 3."
\end{itemize}

\end{methodbox}

\section{Break Down of Evaluation Results}
In this section, we provide the break down evaluation results in Table \ref{tab:cross_model} and Table \ref{tab:distill}. Table \ref{tab:mat} is the results for Mathcheck$^*$, Table \ref{tab:prm} are the results for PRM800K, Table \ref{tab:pb} are the results for ProcessBench.
\begin{table*}[h]
\centering
\caption{Results for MathCheck$^*$}\label{tab:mat}
\begin{tabular}{p{4.5cm}ccc}
\hline
\makecell[l]{Model \\ Method} & Err & Cor & F1 \\
\hline
\makecell[l]{\textit{GPT-4o mini}\\ Greedy Decoding} & 75.0&82.9  & 78.8\\
\makecell[l]{Majority Voting} & 76.2&85.0  & 80.4 \\
\makecell[l]{Multi-Model Debate} &79.5 &80.3  & 79.9 \\
\rowcolor{lightblue}\makecell[l]{Temporal Consistency (Ours)} & 84.7&85.0  & \textbf{84.8} \\
\hline
\makecell[l]{\textit{GPT-4o}\\ Greedy Decoding} &84.5 &90.2  & 87.3\\
\makecell[l]{Majority Voting} &85.1 &93.3  & 89.0\\
\makecell[l]{Multi-Model Debate} &88.4 &93.3  & 90.8\\
\rowcolor{lightblue}\makecell[l]{Temporal Consistency (Ours)} &89.0 &94.8  & \textbf{91.8}\\
\hline
\makecell[l]{\textit{Llama 3.1 8B Instruct}\\ Greedy Decoding} &44.6 &7.8  & 13.3\\
\makecell[l]{Majority Voting} &64.7 &3.1  & 5.9\\
\makecell[l]{Multi-Model Debate} &62.2 &3.6  & 6.8\\
\rowcolor{lightblue}\makecell[l]{Temporal Consistency (Ours)} &55.8 &65.3  & \textbf{60.2}\\
\hline
\makecell[l]{\textit{Mistral 7B Instruct v0.3}\\ Greedy Decoding} &24.6 &28.5  & 26.4\\
\makecell[l]{Majority Voting} &15.9 &76.2  & 26.3\\
\makecell[l]{Multi-Model Debate} &15.7 &79.3  & 26.2\\
\rowcolor{lightblue}\makecell[l]{Temporal Consistency (Ours)} &34.1 &41.5  & \textbf{37.4}\\
\hline
\makecell[l]{\textit{Deepseek-R1-Llama-8B}\\ Greedy Decoding} & 67.6 & 24.4 & 35.9\\
\makecell[l]{Majority Voting} & 79.8 & 22.8 & 35.5\\
\makecell[l]{Multi-Model Debate} & 75.0 & 45.6 & 56.7\\
\rowcolor{lightblue}\makecell[l]{Temporal Consistency (Ours)} & 81.2 & 83.9 & \textbf{82.5}\\
\hline
\makecell[l]{\textit{Deepseek-R1-Qwen-7B}\\ Greedy Decoding} & 77.9 & 95.9 & 86.0\\
\makecell[l]{Majority Voting} & 81.6 & 99.0 & 89.3\\
\makecell[l]{Multi-Model Debate} & 77.3 & 93.8 & 84.8\\
\rowcolor{lightblue}\makecell[l]{Temporal Consistency (Ours)} & 82.0 & 98.4 & \textbf{89.5}\\
\hline
\end{tabular}
\end{table*}
\begin{table*}[h]
\centering
    \centering
\caption{Results for PRM800K}\label{tab:prm}
\begin{tabular}{p{4.5cm}ccc}
\hline
\makecell[l]{Model \\ Method} & Err & Cor & F1 \\
\hline
\makecell[l]{\textit{GPT-4o mini}\\ Greedy Decoding} &27.8 &43.8 & 34.0\\
\makecell[l]{Majority Voting} &31.3 &47.9 & 37.9 \\
\makecell[l]{Multi-Model Debate} &34.4 &42.5 & 38.0 \\
\rowcolor{lightblue}\makecell[l]{Temporal Consistency (Ours)} &34.4 & 45.2& \textbf{39.0} \\
\hline
\makecell[l]{\textit{GPT-4o}\\ Greedy Decoding} &30.4& 65.8& 41.6\\
\makecell[l]{Majority Voting} &30.4 &71.2 & 42.6\\
\makecell[l]{Multi-Model Debate} &41.9 &64.4& 50.7\\
\rowcolor{lightblue}\makecell[l]{Temporal Consistency (Ours)} &39.2 & 75.3& \textbf{51.6}\\
\hline
\makecell[l]{\textit{Llama 3.1 8B Instruct}\\ Greedy Decoding} &10.1 & 1.4& 2.4\\
\makecell[l]{Majority Voting} &18.9 &4.1 & 6.8\\
\makecell[l]{Multi-Model Debate} &23.3 & 1.4& 2.6\\
\rowcolor{lightblue}\makecell[l]{Temporal Consistency (Ours)} &15.0 & 42.5& \textbf{22.1}\\
\hline
\makecell[l]{\textit{Mistral 7B Instruct v0.3}\\ Greedy Decoding} & 11.5& 15.1& 13.0\\
\makecell[l]{Majority Voting} & 6.6&71.2 & 12.1\\
\makecell[l]{Multi-Model Debate} &6.6 & 71.2& 12.1\\
\rowcolor{lightblue}\makecell[l]{Temporal Consistency (Ours)} & 10.6& 17.8& \textbf{13.3}\\
\hline
\makecell[l]{\textit{Deepseek-R1-Llama-8B}\\ Greedy Decoding} & 30.0 & 16.4 & 21.2\\
\makecell[l]{Majority Voting} & 41.0 & 42.5 & 41.7\\
\makecell[l]{Multi-Model Debate} & 42.3 & 52.1 & 46.7\\
\rowcolor{lightblue}\makecell[l]{Temporal Consistency (Ours)} & 39.2 & 69.9 & \textbf{50.2}\\
\hline
\makecell[l]{\textit{Deepseek-R1-Qwen-7B}\\ Greedy Decoding} & 33.9 & 72.6 & 46.2\\
\makecell[l]{Majority Voting} & 41.9 & 80.8 & 55.1\\
\makecell[l]{Multi-Model Debate} & 38.8 & 75.3 & 51.2\\
\rowcolor{lightblue}\makecell[l]{Temporal Consistency (Ours)} & 44.5 & 82.2 & \textbf{57.7}\\
\hline\end{tabular}
\end{table*}

\begin{table*}[h]
\caption{Results for ProcessBench}\label{tab:pb}
\begin{tabularx}{\textwidth}{Xccc@{\hspace{3pt}}ccc@{\hspace{3pt}}ccc@{\hspace{3pt}}ccc}
\hline
\multicolumn{13}{c}{ProcessBench} \\
\hline
\makecell[l]{Model \\ Method} & \multicolumn{3}{c}{GSM8K} & \multicolumn{3}{c}{MATH} & \multicolumn{3}{c}{OlympiadBench} & \multicolumn{3}{c}{Omni-MATH} \\
& Err & Cor & F1 & Err & Cor & F1 & Err & Cor & F1 & Err & Cor & F1 \\\hline
\makecell[l]{\textit{GPT-4o mini}\\ Greedy Decoding} &54.1  &82.9  &65.5  &47.0  & 69.2 &56.0  & 39.0 & 55.2 & 45.7 &35.7  & 58.1 &44.2 \\
\makecell[l]{Majority Voting} & 56.0 & 85.0 & 67.5 & 47.8 & 71.6 & 57.3 & 38.9 & 60.5 & 47.3 &36.1  &58.1  & 44.5 \\
\makecell[l]{Multi-Model Debate} & 63.8 & 80.3 & 71.1 & 52.9 & 64.4 & 58.1 & 42.1 & 49.9 & 45.6 & 40.3 & 47.7 &43.7 \\
\rowcolor{lightblue}\makecell[l]{Temporal Consistency (Ours)} &  63.3& 85.0 & \textbf{72.5} & 51.3 & 74.1 & \textbf{60.7} & 43.1 & 60.8 & \textbf{50.4}  & 41.2 & 61.0 & \textbf{49.2} \\
\hline
\makecell[l]{\textit{GPT-4o}\\ Greedy Decoding} & 70.0 & 90.2 & 78.8 &53.4  & 77.1 &63.1  & 44.8 & 67.0 &53.7 &  46.4& 65.1 & 54.2 \\
\makecell[l]{Majority Voting} & 73.4 & 93.3 & 82.2 & 53.9 & 82.5 & 65.2 & 48.3 &72.8  &58.0  & 49.2 & 71.4 & 58.3 \\
\makecell[l]{Multi-Model Debate} & 77.8 & 93.3 &\textbf{84.8}  & 61.4 & 77.0 & 68.4 & 53.7 &59.5  & 56.4 & 56.1 & 58.9 & 57.5 \\
\rowcolor{lightblue}\makecell[l]{Temporal Consistency (Ours)} & 74.9 & 94.8 & 83.7 & 58.1 & 90.1 &\textbf{70.6}  & 45.8 & 86.7 & \textbf{60.0} &  48.7&86.3  &  \textbf{62.2} \\
\hline
\makecell[l]{\textit{Llama 3.1 8B Instruct}\\ Greedy Decoding} & 23.7 & 7.8 & 11.7 &  16.5&2.5  & 4.3 & 8.3 & 3.2 & 4.7 & 7.8 &3.7  & 5.0 \\
\makecell[l]{Majority Voting} & 41.1 & 3.1 & 5.8 & 30.6 & 1.7 & 3.3 & 19.8 & 4.1 & 6.8 & 25.4 & 2.5 & 4.5 \\
\makecell[l]{Multi-Model Debate} & 45.9 & 3.6 & 6.7 & 37.9 & 3.7 & 6.7 & 30.6 & 2.9 &5.4  & 32.0 & 2.5 &4.6 \\
\rowcolor{lightblue}\makecell[l]{Temporal Consistency (Ours)} & 34.8 & 65.3 & \textbf{45.4} & 28.8 &51.5  & \textbf{36.9} & 23.8 &37.5  &\textbf{29.1}  & 24.6 & 40.7 & \textbf{30.7} \\
\hline
\makecell[l]{\textit{Mistral 7B Instruct v0.3}\\ Greedy Decoding} & 27.1 & 28.5 & \textbf{27.8} &23.7  & 20.9 & \textbf{22.2} & 14.8 & 14.7 & 14.8 & 16.3 & 16.2 & 16.3 \\
\makecell[l]{Majority Voting} & 12.6 & 76.2 & 21.6 & 11.8 & 69.7 & 20.2 & 7.6 & 65.8 & 13.6 & 8.4 & 67.2 & 15.0 \\
\makecell[l]{Multi-Model Debate} & 12.6 & 79.3 & 21.7 &12.0  & 70.2 & 20.4 & 7.3 & 67.0 & 13.1 &8.7  & 66.0 & 15.4 \\
\rowcolor{lightblue}\makecell[l]{Temporal Consistency (Ours)} & 20.8 & 41.5 & 27.7 & 19.4 & 25.9 & 22.1 & 18.0 & 19.8 & \textbf{18.8} & 16.2 & 31.5 & \textbf{21.4} \\
\hline
\makecell[l]{\textit{Deepseek-R1-Llama-8B}\\ Greedy Decoding} & 44.9 & 24.4 & 31.6 & 45.5 & 24.1 & 31.5 & 35.1 & 24.8 & 29.0 & 31.2 & 20.7 & 24.9 \\
\makecell[l]{Majority Voting} & 49.3 & 22.8 & 31.2 & 67.5 & 50.0 & 57.4 & 57.3 & 58.7 & 58.0 & 51.8 & 46.5 & 49.0 \\
\makecell[l]{Multi-Model Debate} & 51.7 & 45.6 & 48.5 & 64.5 & 63.8 & 64.1 & 56.1 & 71.1 & 62.7 & 49.9 & 61.0 & 54.9 \\
\rowcolor{lightblue}\makecell[l]{Temporal Consistency (Ours)} & 56.5 & 83.9 & \textbf{67.6} & 67.0 & 79.6 & \textbf{72.7} & 57.0 & 78.5 & \textbf{66.1} & 53.1 & 75.1 & \textbf{62.2} \\
\hline
\makecell[l]{\textit{Deepseek-R1-Qwen-7B}\\ Greedy Decoding} & 52.2 & 95.9 & 67.6 & 50.5 & 80.0 & 61.9 & 39.0 & 64.6 & 48.7 & 29.6 & 66.0 & 40.9 \\
\makecell[l]{Majority Voting} & 57.5 & 99.0 & 72.7 & 64.3 & 88.4 & 74.5 & 48.1 & 81.7 & 60.6 & 39.0 & 75.5 & 51.4 \\
\makecell[l]{Multi-Model Debate} & 58.0 & 93.8 & 71.7 & 59.8 & 84.7 & 70.1 & 45.8 & 71.1 & 55.7 & 37.7 & 71.4 & 49.3 \\
\rowcolor{lightblue}\makecell[l]{Temporal Consistency (Ours)} & 62.8 & 98.4 & \textbf{76.7} & 69.5 & 94.3 & \textbf{80.1} & 54.5 & 90.6 & \textbf{68.0} & 46.1 & 86.7 & \textbf{60.2} \\
\hline
\end{tabularx}
\end{table*}

\end{document}